\documentclass[a4paper,twoside]{article}

\usepackage{epsfig}
\usepackage{subcaption}
\usepackage{calc}
\usepackage{amssymb}
\usepackage{amstext}
\usepackage{amsmath}
\usepackage{amsthm}
\usepackage{multicol}
\usepackage{pslatex}
\usepackage{apalike}
\usepackage{algorithm2e}
\usepackage[bottom]{footmisc}
\usepackage{SCITEPRESS}     

\begin{document}

\title{RetailKLIP : Finetuning OpenCLIP backbone using metric learning on a single GPU for Zero-shot retail product image classification}

\author{\authorname{Muktabh Mayank Srivastava }
\affiliation{ParallelDots, Inc.}
\email{muktabh@paralleldots.com }}


\abstract{Retail product or packaged grocery goods images need to classified in various computer vision applications like self checkout stores, supply chain automation and retail execution evaluation. Previous works explore ways to finetune deep models for this purpose. But because of the fact that finetuning a large model or even linear layer for a pretrained backbone requires to run at least a few epochs of gradient descent for every new retail product added in classification range, frequent retrainings are needed in a real world scenario. In this work, we propose finetuning the vision encoder of a CLIP model in a way that its embeddings can be easily used for nearest neighbor based classification, while also getting accuracy close to or exceeding full finetuning. A nearest neighbor based classifier needs no incremental training for new products, thus saving resources and wait time.}

\onecolumn \maketitle \normalsize \setcounter{footnote}{0} \vfill

\section{\uppercase{Introduction}}
\label{sec:introduction}

Recognizing Retail Product or packaged grocery goods in images or videos can help to solve multiple problems in supermarket floors and supply chain hubs. Enabling self checkout stores, measuring retail execution to evaluate merchandising activities and automatic slotting of products during fulfillment are some of the applications. Retail Product Image Classification has generally been treated as one-shot or few-shot classification problem, because, unlike say a class like dog or cat, the variance between different images of a retail product is much lesser and just involves different occlusions, blurring and lighting changes. Finetuning or Transfer learning in Deep Neural Networks, often Convolutional Neural Networks, has been used in existing literature for this purpose. However, even in such cases, previous classification methods require gradient descent to be run to train classifiers for the full neural network backbone or the last linear layer to get best results. Because grocery products have very frequent new launches, package redesigns and offer tags, this involves finetuning quite often. This process requires usage of computing resources to finetune models being used to classify 100s or 1000s of products, again, even when say 2 images of a newly launched product are discovered. This process not just makes maintaining real world Retail Product Classification models costlier, it also creates a "blind spot" time, where for the time being a new model is being finetuned to add new products as classes, the model cannot recognize these new products.

CLIP is a way to train an image encoder and a text encoder in parallel such that the embedding of an image and its text description are close in a common space \cite{radford2021learning} . OpenCLIP is an implementation of CLIP available under a permissive license \cite{ilharco_gabriel_2021_5143773} . Because of large number of training examples and text descriptions providing descriptive annotations for objects, the CLIP image encoder has been used for many classification problems with a linear layer or even nearest neighbors. While embeddings from CLIP encoder are much better than any other pretrained weights to use in zero shot setting and can be used as good baselines, finetuning them for a specific domain is always desirable to make the zero shot classification more accurate. However, CLIP has generally proven to be hard to finetune. We propose an end to end pipeline to finetune a CLIP backbone for zero shot Retail Product image classification, identifying and addressing different concerns.

Our work has two components : 1. Creating a large dataset which can be used for finetuning a CLIP backbone for retail product images. 2. Proposing a learning rate strategy, class balancing approach and other finetuning components which solve for erratic finetuning of CLIP backbones and address imbalanced large retail datasets. 

Our constraint in this work is that we have to limit our work on one GPU and so cannot train any models which go beyond a single GPU both for Zero Shot classification or any baselines.

\section{\uppercase{Related Work}}

Retail Product Image Recognition has been studied using multiple techniques. Older works use feature descriptors like SIFT and BRISK \cite{6126542,SIFT} to recognize retail products. More recently, tricks to finetune Convolutional Neural Networks \cite{Srivastava,peng2020rp2k}  to perform well in the task were proposed. However, given finetuning the entire backbone is costly, recent works propose training just a linear layer to learn classify representations given by a backbone trained by contrastive supervised/semi-supervised methodology on relevant datasets \cite{srivastava2022using} . There have also been works where GAN-like backbones have been used to recognize retail products using information retrieval techniques \cite{Tonioni2019}.

More recently, there has been a trend to use LVMs (Large Vision Models) as a backbone instead of traditional ImageNet pretrained backbones for transfer learning in Computer Vision. CLIP is one method of training these LVMs where Billion plus sized datasets of images and their text descriptions are trained to learn image embedding and text embedding of description of the image such that they lie close on a common space \cite{radford2021learning} . Because these datasets are huge and text descriptions have more details than just a classification dataset, this training mechanism yields excellent image encoders. CLIP Image encoders have shown great results on zero shot and linear layer only training tasks on internet images. A CLIP image encoder for example can be used to get excellent results on ImageNet, by just simply comparing the embeddings or training a linear layer on top of embeddings generated. OpenCLIP is an implementation of CLIP with permissive license, that we use in our work \cite{ilharco_gabriel_2021_5143773}.

CLIP image encoder backbone however is considered hard to finetune for related but different domains. There have been some publications which give some hints about finetuning \cite{dong2022clip,kumar2022fine}. Their work inspires our finetuning technique. The image encoder finetuned has a Vision Transformer architecture (specifically, a large variant of Vision transformer or ViT-L) \cite{dosovitskiy2020image}.

Deep Metric Learning \cite{Musgrave2020PyTorchML} techniques can be used to learn encoders which generate discriminative embeddings. Unlike softmax based classification, metric learning approaches generalize better to openset recognition \cite{deng2019arcface}. ArcFace loss, a metric learning loss function, used generally in long tailed facial recognition tasks is used to finetune CLIP \cite{deng2019arcface}. To our knowledge, this is the first time metric learning has been used to finetune a LVM to a new domain.

Given that retail datasets are very imabalanced is another reason we use ArcFace loss to finetune the CLIP model to the retail domain. Previous works have shown that plain ArcFace handles imbalanced datasets better than softmax loss even when aided with data balancing techniques \cite{9150576} . We use ArcFace loss with a custom data balancing technique for our final results. 

\section{\uppercase{Datasets}}
We would like to list the datasets used for the experiments in our work. To finetune CLIP image encoder to retail product image recognition, we use an inhouse (not publicly available) dataset called RP6K with 6500 retail products. This has over 1 Million images of retail products with their class tags. Just like real world, the number of images of these products in RP6K is long-tailed with some products having upto 4000 images, while many having lesser than 10 images. We finetune CLIP image encoder on this dataset using a novel mixture of techniques which yields a model that can be used for Zero shot classification on other datasets. \label{sec:RP6K}

Grozi-120 is the first dataset used for evaluation \cite{Merler-vitro}. It is a one shot classification dataset with one train image per product, which is packshot like, while test set images are from real world. It has 120 products.

CAPG-GP \cite{Geng-acm} is another one shot dataset, but with fine grained products. However, the train and test images both appear to be from similar domain here.

We have explicitly made efforts to make sure none of the products in CAPG-GP and Grozi-120 are present in RP6K.

RP2K \cite{peng2020rp2k} is a dataset with 2388 retail products, with substantial number of both train and test images for each product. However, we still use the dataset in a few shot setting like Grozi120 and CAPG-GP. That is, while the entire test set will be used to test the performance of Zero Shot classification, only one image per class from train set will be used to calculate class representations for classification. 

\section{\uppercase{Method}}
Our work involves finetuning OpenCLIP image encoder on a retail product image dataset RP6K such that resultant image encoder can yield embeddings of product images useful for zero shot retail product image classification on other datasets. The image encoder after finetuning with RP6K is referred to as RetailKLIP. For zero shot classification, one image per product is taken and augmented and then passed to RetailKLIP to get embeddings for that product. This process is repeated for all products to create a set of products' embeddings. To test an unseen image, its image embedding created using RetailKLIP is compared to product embeddings created earlier and the product with closest embedding is considered to be the output of RetailKLIP. 

\begin{figure*}[!h]
  \centering
   {\epsfig{file = 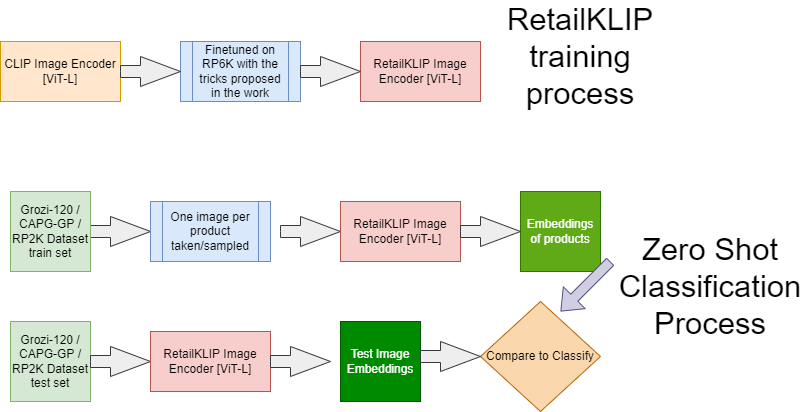, width = 11cm}}
  \caption{RetailKLIP is trained on RP6K. Its then evaluated for Zero Shot classification on Grozi-120, CAPG-GP and RP2K}
  \label{fig:process}
 \end{figure*}

\subsection{Finetuning OpenCLIP to RetailKLIP}
We use the ViT-L OpenCLIP pretrained model to obtain RetailKLIP through finetuning. Specific Architecture used is ViT-L/14 trained on LAION-2B available on OpenCLIP's Github \cite{ilharco_gabriel_2021_5143773}. We take the trick of using AdamW optimizer to finetune CLIP from \cite{kumar2022fine}. We also use differing learning rate with depth like in \cite{dong2022clip}, but instead of layerwise rate decay, we use block wise rate decay. That is, within each block of ViT there is a common learning rate, but learning rate varies across blocks. RP6K is a very imbalanced dataset, hence we use ArcFace loss to finetune OpenCLIP. Specifically, ArcFace \cite{deng2019arcface} implementation of PyTorch Metric Learning library \cite{Musgrave2020PyTorchML} is used for the task. ArcFace is considered good for openset recognition and we use it because while the finetuning is being done on RP6K dataset, the model is used for Zero Shot classification on CAPG-GP, Grozi-120 and RP2K datasets, making it an openset task. ArcFace seems to also hold itself better against softmax on datasets with a long tail distribution according to previous work, infact just vanilla ArcFace finetuning is better than softmax finetuning with data balancing tricks \cite{9150576}.

\subsubsection{Techniques used for finetuning OpenCLIP on RP6K}
CLIP models have proven to be hard to finetune. However, \cite{kumar2022fine} have proposed methods that can be used to finetune the CLIP vision-backbone for Image Recognition and other Computer Vision tasks. We use AdamW optimizer as is suggested. ArcFace \cite{deng2019arcface} is used as loss function while finetuning.

\subsubsection{Block wise Learning Rate Decay Trick}
We also find like in the case of \cite{dong2022clip} that using a common learning rate for the entire model performs at much lower levels than using differing learning rates across the backbone. However, we replace the LLRD (Layerwise Learning Rate Decay) strategy of the work with Blockwise learning rate decay. LLRD starts with a learning rate for the top layer (classification head in case of cited work) and decreases learning rate for each layer by a constant multiplicative decay.
In our work, we use a blockwise multiplicative decay strategy. The last ViT-L block (top block) starts with learning rate of 2e-4 and each previous block having a decreasing learning rate by a factor of 0.7. The blocks used here are blocks of transformers and other layers labelled in OpenCLIP's code implementation of ViT.

\subsubsection{Data Balancing}
While ArcFace gives good results even without any data balancing while training, we use a custom data balancing approach to finetune on a steep imbalanced dataset like RP6K. While finetuning of RP6K, we create a held out validation set from RP6K dataset with equal number of test images from each product. Average accuracy on this validation set is used to determine optimal data balancing. \cite{9150576} takes 2 concepts into account while data balancing, depth, which is number of datapoints per class in train set and breadth, that is number of classes. We find that keeping breadth to maximum possible gets best average accuracy on the mentioned validation set. We test many values of depth to find the value which gets best avg accuracy on RP6K validation set. So maximum breadth and determined value of depth is used for finetuning. Classes are oversampled or undersampled accordingly with various augmentation tricks to get the balanced dataset.

\subsection{Zero Shot classification using RetailKLIP}
To use RetailKLIP as a Zero Shot classifier on a new dataset, one doesn't need to finetune RetailKLIP. One image per product is taken from trainset and its augmentations are created. These augmentations of a product image are passed through RetailKLIP to get a set of embeddings for the product. Once we have embeddings for all products, we are ready to classify. An image of the test dataset is passed through RetailKLIP to get its embedding and compare it with existing products' embeddings. The nearest embedding from train set is taken to be the result of classification. This is k-nearest-neighbors classification process with k=1.

\section{\uppercase{Results}}
For Grozi-120 \cite{Merler-vitro} and CAPG-GP \cite{Geng-acm}, we compare the accuracy of RetailKLIP Zero Shot classification with full finetuning of a ResNext-WSL model \cite{Mahajan}. We also compare it with tricks like using LCA layer and maxent loss while full finetuning ResNext-WSL to increase accuracy from \cite{Srivastava}. These methods involve full finetuning of the Convnet backbone making the process slow and compute intensive. We also compare it with accuracy of using a pretrained semi supervised backbone with trainable linear layers \cite{srivastava2022using}. While training just a linear layer is also quite fast, zero shot classification with RetailKLIP needs no training at all. The basis for comparing ResNext-WSL with ViT-L despite difference in number of parameters is because these are the largest models than can fit on a single Nvidia 1080Ti GPU for training. 

\begin{table}[h]
\vspace{-0.2cm}
\caption{Results of various Models on CAPG-GP Dataset which can be trained on a single GPU. First two are methods full finetuning a ResNext-WSL backbone, the third is using a semi supervised pretrained backbone and training a Linear Layer. The fourth is zero shot method with RetailKLIP neeeding no training. }\label{tab:results1} \centering
\begin{tabular}{|p{3.5cm}|c|}
  \hline
  Model Name & Accuracy [CAPG-GP] \\
  \hline
  ResNext-WSL (full finetuning)  & 84.1\%   \\
  \hline
  ResNext-WSL+LCA layer+MaxEnt Loss (full finetuning)  & 92.2\% \\
  \hline
  Pretrained Semi Supervised ResNext-WSL backbone + linear layer training  & 87.6\% \\
  \hline
  RetailKLIP (Zero Shot) & 88.6\% \\
  \hline
\end{tabular}
\end{table}

\begin{table}[h]
\vspace{-0.2cm}
\caption{Results of various Models on Grozi-120 Dataset  which can be trained on a single GPU. First two are methods full finetuning a ResNext-WSL backbone, the third is using a semi supervised pretrained backbone and training a Linear Layer. The fourth is zero shot method with RetailKLIP neeeding no training. }\label{tab:results2} \centering
\begin{tabular}{|p{3.5cm}|c|}
  \hline
  Model Name & Accuracy[Grozi-120] \\
  \hline
  ResNext-WSL (full finetuning)  & 60.4\%   \\
  \hline
  ResNext-WSL + LCA layer + MaxEnt Loss (full finetuning)  & 72.3\% \\
  \hline
  Pretrained Semi Supervised ResNext-WSL backbone + linear layer training & 76.19\% \\
  \hline
  RetailKLIP (Zero Shot) & 82.8 \% \\
  \hline
\end{tabular}
\end{table}

As we can see, RetailKLIP gives competitive results to full finetuning or linear layers trained on semi supervised trained backbone.

\subsection{RP2K results}
RP2K dataset paper proposes full finetuning ResNet and other backbones on their dataset one category at a time to get upto 95\% accuracy in some categories of their dataset. For some other harder categories, the accuracy can be lower than 90\%. However, due to language barrier, our team has been unable to separate out categories and thus we have to use models on all categories combined, so this makes the problem harder for RetailKLIP. Also RetailKLIP takes just one image from train set to classify test images unlike full finetuning which uses the ample amount of images available. The accuracy for Zero Shot classification on RP2K is 87.7\%. This is close to the accuracy full finetuning approach of \cite{peng2020rp2k} gets on the hardest categories.

\section{\uppercase{Discussion}}
Our work proposes a method to create a Zero shot classifier for Retail Product images on a single GPU by finetuning OpenCLIP. The accuracy is competitive or even sometimes better than full finetuning large Convnet backbones on the same GPU. This enables real world retail computer vision systems to quickly integrate new products into their range and avoid resource intensive trainings multiple times.

\bibliographystyle{apalike}
{\small
\bibliography{example}}

\end{document}